\documentclass[preprint,12pt]{elsarticle}

\usepackage{amssymb}
\usepackage{amsmath}

\usepackage{amsmath,amssymb,amsfonts}
\usepackage{algorithmic}
\usepackage{graphicx}
\usepackage{algorithm,algorithmic}
\usepackage{hyperref}

\usepackage{textcomp}

\usepackage{amsmath}
\usepackage{amssymb}
\usepackage{algorithm}
\usepackage{float}
\usepackage{booktabs}
\usepackage{tikz}
\usepackage{booktabs} 
\usepackage{multirow} 
\usetikzlibrary{positioning,arrows.meta}

\journal{Robotics and Autonomous Systems}

\begin{document}

\begin{frontmatter}



\title{A Prior Knowledge-based Deep Reinforcement Learning Method Enhancing Autonomous Guidewire Navigation Training Effect for Endovascular Surgeries}


\author[label1]{Chong Yang} 
\author[label1,label2]{Shuxiang Guo\corref{cor1}}
\author[label1]{Weihao Wu}
\author[label1]{Haoyu Xie}

\cortext[cor1]{Corresponding author. E-mail: guo.shuxiang@sustech.edu.cn} 

\affiliation[label1]{organization={The Department of Electronic and Electrical Engineering, Southern University of Science and Technology},
            city={Shenzhen},
            state={Guangdong},
            country={China}}

\affiliation[label2]{organization={The Aerospace Center Hospital, School of Life Science, and the Key Laboratory of Convergence Medical Engineering System and Healthcare Technology, Ministry of Industry and Information Technology,
Beijing Institute of Technology},
            city={Beijing},
            country={China}}

\begin{abstract}
This research aims to address the critical challenges of applying Deep Reinforcement Learning (DRL) to autonomous guidewire navigation in vascular interventional surgery, namely the sample inefficiency of training and the anatomically misleading signals from conventional reward metrics. A novel DRL framework that strategically integrates anatomical prior knowledge is presented. A Dijkstra-based geodesic distance metric is employed in place of the Euclidean distance to define a reward signal that better reflects vascular anatomy. In addition, an efficient discretized, stacked state-space representation is introduced to substantially reduce dimensionality and alleviate partial observability. These components are integrated within a Deep Q-Network (DQN) architecture featuring an action-differentiated reward scheme. Simulations are conducted on a hepatic venous model and the results demonstrated that the proposed method substantially accelerated training convergence and effectively suppressed oscillatory behaviors, learning more direct and optimal trajectories. 
\end{abstract}



\end{frontmatter}

\section{Introduction}

The rising global incidence of cardiovascular and cerebrovascular diseases has positioned minimally invasive endovascular interventions as a critical therapeutic paradigm\cite{dreyfus2024}\cite{bao2024}. These procedures are favored for their reduced physical trauma, shorter hospital stays, and improved patient outcomes \cite{yan2025a}\cite{yan2025b}. However, a significant limitation persists: the substantial radiation exposure clinicians face during fluoroscopy-guided interventions\cite{bao2022}\cite{cao2024}. Robotic-assisted platforms have emerged as a viable solution to this occupational hazard \cite{guo2019a}\cite{yan2023a}, enabling surgeons to conduct procedures from a safe distance and offering the potential to bridge geographic disparities in access to specialized care \cite{lyu2024a}\cite{lyu2023flexible}.

Despite these advantages, contemporary robotic systems primarily function as supportive tools, with limited decision-making autonomy\cite{li2024robotic}. The success of complex navigation tasks, such as manipulate a guidewire through intricate and tortuous vascular networks, remains heavily reliant on the operator's expertise and real-time judgment\cite{li2024b}. This dependency can lead to prolonged procedure times, increased patient risk due to extended fluoroscopy exposure, and a steep learning curve for novice surgeons\cite{pore2023autonomous}. The automation of this navigation process holds the promise of alleviating surgeon burden, enhancing procedural consistency, and improving overall safety \cite{mei2025}.

The integration of artificial intelligence (AI), particularly Reinforcement Learning (RL), represents a paradigm shift in the development of autonomous surgical robotic systems\cite{sun2025}. The fundamental strength of RL lies in its framework for learning optimal control policies through environmental interaction, making it exceptionally well-suited for mastering complex navigation strategies in uncertain and dynamic environments. This capability is crucial for endovascular interventions, where robotic systems must adapt to intricate vascular anatomies and real-time physiological variations.

However, training effective RL agents for such delicate tasks presents formidable challenges. A primary obstacle is the sample inefficiency of these algorithms, as RL agents typically require an enormous number of interactions to acquire effective policies. Furthermore, the high-dimensional nature of the state space contributes to the curse of dimensionality, leading to slow convergence and demanding extensive training iterations. These constraints have driven the widespread adoption of high-fidelity simulation environments, such as the SOFA, which provides a physically accurate and safe virtual platform for training without risking patient safety. SOFA enables realistic modeling of complex interactions between flexible instruments and deformable vascular structures, thereby offering a scalable foundation for data acquisition and policy refinement \cite{yao2025b}.

Within such simulated environments, a spectrum of DRL algorithms has been investigated. The Soft Actor-Critic (SAC) algorithm has achieved robust autonomous navigation across multiple benchmark scenarios with high success rates\cite{karstensen2025}. To further address the critical issue of sample efficiency, advanced algorithms have been developed to learn effective policies from sparse environmental feedback. A notable exemplar is Deep Deterministic Policy Gradient with Hindsight Experience Replay (DDPG-HER) \cite{karstensen2022}, which has been successfully applied to train neural network controllers for autonomous guidewire navigation in virtual environments, such as the venous system of a porcine liver, without relying on human demonstration data.

Li et al.\cite{li2024} proposed a CASOG (Conservative Actor–Critic with Smooth Gradient) offline reinforcement learning algorithm, which utilizes the guidewire operation data collected from the real environment to conduct offline training of guidewire operation skills in the vascular intervention robot system. The experiment proved that it is possible to successfully deliver the guidewire to the target position. Additionally, structured training curricula using algorithms like Proximal Policy Optimization (PPO) with multi-stage training have been employed to progressively master navigation tasks, achieving successful validation in real-world scenarios \cite{mei2025}.                                                                                                                                                                                                                                                                                                                   

Despite these advancements, a fundamental limitation persists in the design of reward functions, where existing approaches often rely on simplistic metrics like Euclidean distance. In tortuous vasculature, such metrics provide anatomically misleading signals \cite{song2022}, resulting in oscillatory behaviors and convergence to local optima that hinder the learning of robust navigation policies. To address this, we propose a prior knowledge-integrated Deep Reinforcement Learning framework featuring a novel reward structure grounded in anatomical prior knowledge. Our key innovation utilizes a precomputed distance field—calculated via Dijkstra’s algorithm on a 2D projection of the vascular model—to replace Euclidean distance, thereby providing a consistent, monotonic learning signal. Complementing this, we introduce a discretized, stacked state space representation integrated within a Deep Q-Network (DQN) architecture featuring an action-differentiated reward scheme; this design dramatically reduces dimensionality, mitigates partial observability, and discourages oscillatory actions. Through extensive comparative experiments in SOFA, we demonstrate that this synergistic combination of anatomical priors and an efficient learning architecture is paramount for robust autonomous navigation. Our approach not only achieves a high success rate but also significantly accelerates training convergence and yields more efficient navigation policies compared to conventional benchmarks.

\begin{figure*}[htbp]
    \centerline{\includegraphics[width=1\linewidth]{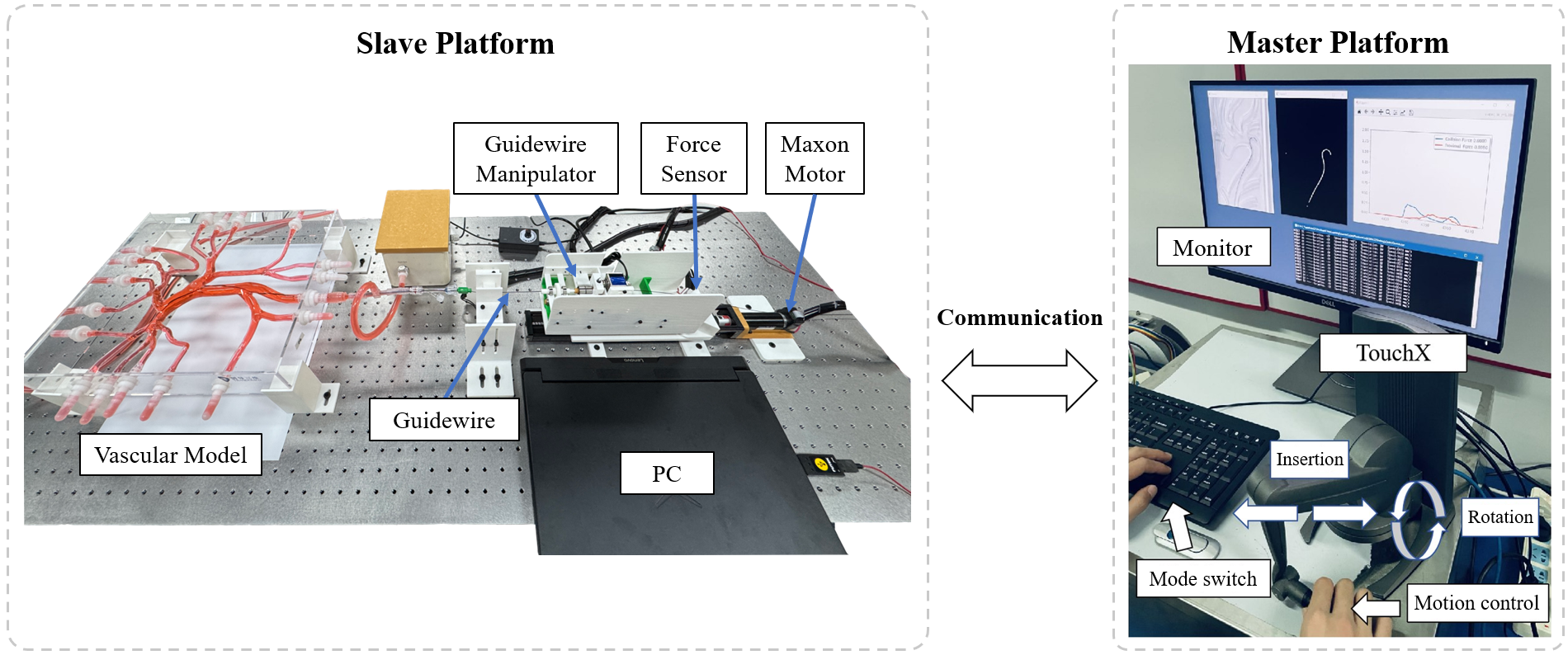}}
    \caption{A master-slave operated endovascular surgical robot.}
    \label{robot}
\end{figure*}

\section{Autonomous Guidewire Navigation Method}

\subsection{Guidewire Navigation Task}

Precise and secure insertion of the guidewire to the target lesion is a prerequisite for subsequent therapeutic maneuvers. The guidewire serves as the rail along which catheters and other devices are delivered. In past work, We have already designed a master-slave operated vascular interventional surgical robot, as shown in Fig.~\ref{robot}. 
However, the intricate arborization of vascular renders this task nontrivial; for inexperienced operators, repeated probing is usually required, prolonging fluoroscopy time and markedly increasing the patient's cumulative radiation dose, thereby elevating procedural risk. Automating this step under stringent safety constraints can improve success rates, shorten procedure duration, and substantially reduce operator burden.

\begin{figure*}[htbp]
    \centerline{\includegraphics[width=1\linewidth]{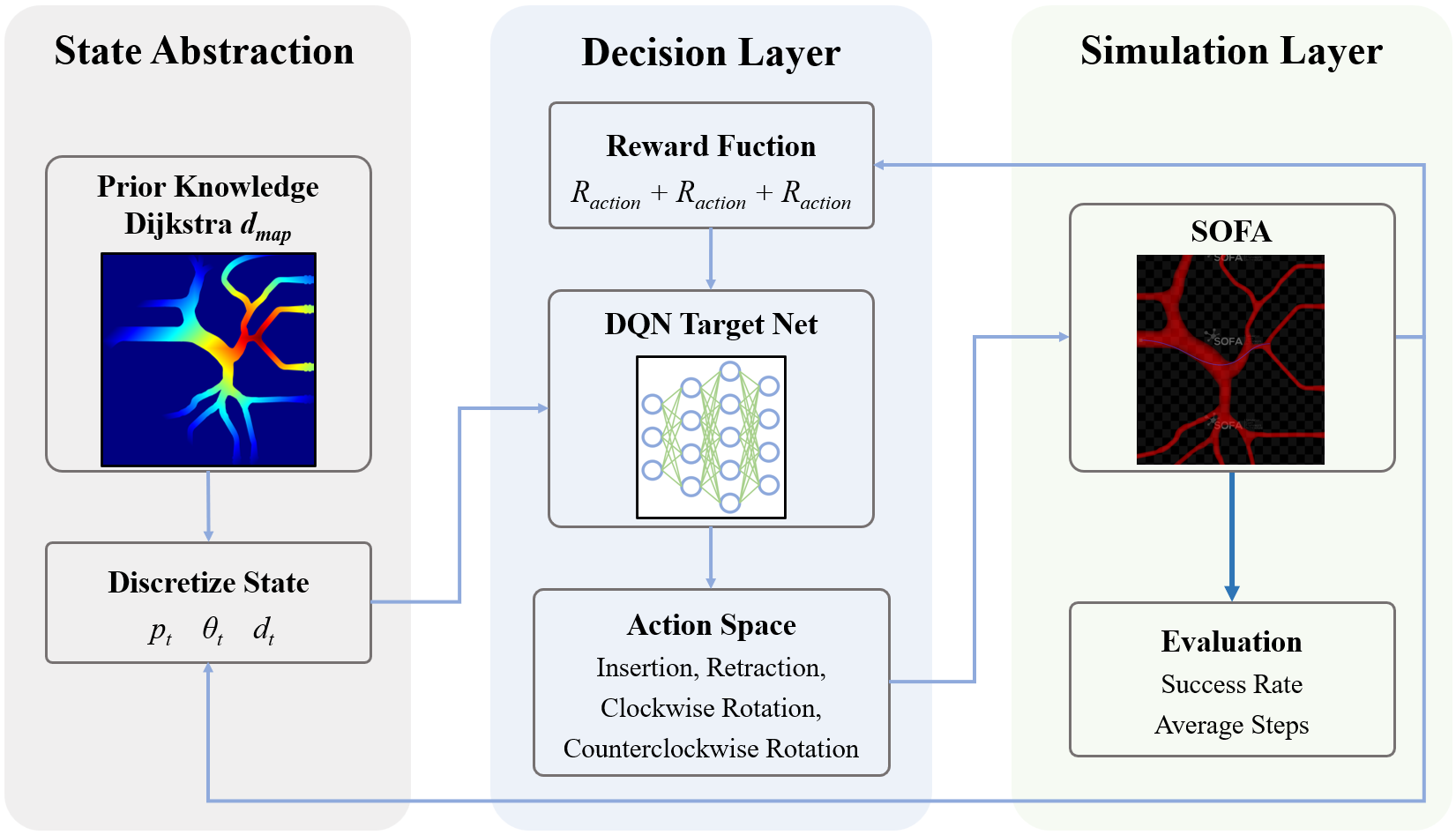}}
    \caption{The overall system pipeline of the proposed prior knowledge-integrated Deep Reinforcement Learning  framework for autonomous guidewire navigation. The architecture consists of three core modules: the State Abstraction layer incorporates anatomical prior knowledge by precomputing a Dijkstra distance field ($d_{map}$) and transforms continuous variables (shortest path $p_t$, orientation $\theta_t$, and insertion depth $d_t$) into a discretized and stacked state representation to reduce dimensionality and mitigate partial observability; the Decision Layer utilizes a Deep Q-Network to select optimal discrete actions based on reward function; and the Simulation Layer executes these commands within the high-fidelity SOFA environment to model complex physical interactions and evaluate navigation performance.}
    \label{overall_system}
\end{figure*}

The automation of this procedure presents unique challenges, including vessel tortuosity, bifurcation selection, and safety constraints. We formulate this as a sequential decision-making problem.
We propose a deep reinforcement learning algorithm that integrates prior knowledge for achieving autonomous navigation in vascular interventional surgeries, the overall system pipeline of the prior knowledge integrated DRL guidewire navigation is as shown in Fig.~\ref{overall_system}. This significantly accelerated the training speed of the agent, reducing the training time and cost. Specifically, we calculated the path lengths from each point in the vascular path to the target point using the Dijkstra's algorithm as prior knowledge, and then incorporated it into the state representation of the agent.

\subsection{System Architecture and Simulation Environment}
The reinforcement-learning training environment is built upon the open-source multi-physics simulation framework SOFA. Leveraging real-time finite-element and constraint solvers, SOFA accurately captures the nonlinear deformation, contact, and frictional interactions between a guidewire and a complex vascular network.

\subsection{Deep Reinforcement Learning Framework}
Intravascular interventional surgery requires the guidewire to complete high-precision and low-damage navigation in complex and tortuous vascular trees. Due to significant individual differences in vascular morphology, traditional control strategies based on rules or path planning are difficult to adapt to anatomical variations. This paper proposed and fully implemented an autonomous guidewire navigation framework based on DRL, modeling the surgical task as a Markov decision process (MDP), and completing training and verification in the SOFA.

To solve this optimization problem, we employ the DQN algorithm, which achieves stable and efficient learning by minimizing the temporal-difference (TD) error. The DQN objective is defined as

\begin{equation}
L(\theta) = \mathbb{E}_{(s,a,r,s')\sim\mathcal{D}}\Bigl[\Bigl(r+\gamma\max_{a'}Q_{\bar{\theta}}(s',a')-Q_{\theta}(s,a)\Bigr)^{2}\Bigr],
\end{equation}
where $\mathcal{D}$ denotes a mini-batch uniformly sampled from the experience-replay buffer, $\theta$ represents the parameters of the current Q-network, $\bar{\theta}$ are the parameters of the target network (copied from $\theta$ every $C$ steps and kept fixed during training), and $\gamma$ is the discount factor. The expectation $\mathbb{E}$ averages the squared TD error, driving the predicted Q-values toward the target values and implicitly improving the policy.

The Q-network is parameterized as a multi-layer perceptron $f_{\theta}(\cdot)$ with ReLU activations, producing Q-values for each discrete action:

\begin{equation}
Q_{\theta}(s,a)=f_{\theta}(s)[a].
\end{equation}

Since DQN is inherently designed for discrete action spaces, we quantize the guidewire control commands into four atomic actions. The agent then selects the action with the highest Q-value:

\begin{equation}
a_{t}=\arg\max_{a}Q_{\theta}(s_{t},a).
\end{equation}

\subsection{State Space Representation}

The state vector $s_t \in \mathcal{S}$ is designed to capture both spatial and temporal information for effective navigation decision-making. To address the partial observability of the environment, we employ a stacked state representation incorporating historical information.

\subsubsection{Basic State Components}
\begin{itemize}
    \item $p_t$: Shortest-path length from the guidewire tip to the target, computed using Dijkstra's algorithm on the vessel projection graph.
    \item $\theta_t$: Orientation angle of the guidewire tip, encoded as a scalar value derived from quaternion representation.
    \item $d_t$: Accumulated insertion length from the vascular entry point. Owing to multiple bifurcations, relying solely on \(p\) can yield identical tip-to-target distances for distinct anatomical configurations; incorporating \(d\) disambiguates these cases and guarantees a unique state identifier.
\end{itemize}

\subsubsection{State Space Discretization Strategy}
The continuous nature of guidewire navigation presents significant challenges for reinforcement learning, including high-dimensional state spaces and difficult function approximation. Our discretization approach addresses these challenges through principled dimension reduction.

The discretization process can be viewed as a function approximation problem where we project the continuous state space onto a discrete lattice. For a Lipschitz-continuous value function $Q(s,a)$, the discretization error is bounded by:
\begin{equation}
|Q(s,a) - Q(\hat{s},a)| \leq L \cdot \|s - \hat{s}\|
\end{equation}
where $L$ is the Lipschitz constant and $\hat{s}$ is the discretized state. Our binning strategy ensures that the maximum discretization error remains within acceptable limits while dramatically reducing computational complexity.

Each component is uniformly discretized to reduce state space complexity:
\begin{align*}
    p_t & \rightarrow p_t^{\text{disc}} \in \{0, 1, \ldots, N_p-1\} \\
    \theta_t & \rightarrow \theta_t^{\text{disc}} \in \{0, 1, \ldots, N_\theta-1\} \\
    d_t & \rightarrow d_t^{\text{disc}} \in \{0, 1, \ldots, N_d-1\}
\end{align*}

The number of bins for each state component is determined through sensitivity analysis to provide sufficient resolution for precise navigation while maintaining computational tractability.

\subsubsection{Stacked State Representation}
The guidewire navigation task exhibits characteristics of a Partially Observable Markov Decision Process (POMDP). A single time step's state descriptor $s_t = [p_t, \theta_t, d_t]$, while compact, provides only a static snapshot of the system. This instantaneous representation is insufficient for uniquely determining the system's true dynamics and can lead to ambiguous state interpretations. 

To mitigate this partial observability and ensure accurate state identification, we employ a temporal state stacking strategy. The augmented state is constructed by concatenating observations from three consecutive time steps:

\begin{equation}
s_t^{\text{stacked}} = [s_{t-2}, s_{t-1}, s_t]
\end{equation}

where each $s_i = [p_i^{\text{disc}}, \theta_i^{\text{disc}}, d_i^{\text{disc}}]$ represents the discretized state vector at time step $i$. 

\subsection{Action Space Representation}
The action space A denotes the set of executable actions available to the robotic system at each decision-making step. To reduce training complexity while ensuring successful autonomous navigation, we have simplified the guidewire manipulation into four fundamental actions: 2mm insertion, 2mm retraction, 5° clockwise rotation, and 5° counterclockwise rotation.

\subsection{Reward Function Design}

The design of the reward function is a critical determinant of success for Deep Reinforcement Learning (DRL) agents, especially in safety-critical applications like endovascular navigation. Conventional approaches that rely on Euclidean distance provide a geometrically simple but anatomically misleading signal in complex vascular environments, often leading to suboptimal policies prone to oscillations, inefficiency, and navigational failures at bifurcations. To address these fundamental limitations, we introduce a novel reward structure deeply integrated with anatomical prior knowledge. Our fundamental contribution is the introduction of a computationally efficient yet anatomically-grounded 2D Projected Distance Field, computed using Dijkstra's algorithm, as the core metric for reward calculation. Our design philosophy centers on providing the agent with anatomically-meaningful and action-aware feedback to promote the emergence of robust and efficient policies that respect the topological constraints of the vascular network.

\subsubsection{Distance Field Construction}

The choice of distance metric is critical, as it fundamentally shapes the learning signal for the reinforcement learning agent. Our approach is motivated by the significant limitations of using Euclidean distance in complex vascular environments. The Euclidean metric, defined as the straight-line distance between two points, provides a geometrically simple but anatomically misleading signal. As a guidewire navigates through tortuous paths and bifurcations, the Euclidean distance to the target can paradoxically increase even when the agent is advancing along the correct anatomical path. This non-monotonic behavior creates spurious local optima, often rewarding the agent for entering incorrect branches that happen to align spatially with the target and penalizing correct navigation through curved segments. This misleading guidance severely undermines the stability and efficacy of the learning process.

To overcome this fundamental issue while maintaining computational tractability for intensive DRL training, we strategically employ a 2D projected distance field as an intermediate object of study. This decision is based on the rationale that successfully solving the navigation problem in a computationally efficient 2D domain is a necessary and valuable first step before tackling the full complexity of 3D space. The core of our method utilizes Dijkstra's algorithm propagated through the 4-connected neighborhood on a 2D binary image of the vasculature. This method computes the shortest path distance within the navigable region, rather than the straight-line distance.

The key advantage of our Dijkstra-based distance field over the Euclidean metric is its path-aware monotonicity. As the agent moves along any viable path toward the target, the computed distance is guaranteed to decrease monotonically. This provides a consistent and reliable learning gradient, which is essential for stable policy optimization and established a robust foundation for training the DRL agent.

\begin{algorithm}[h]
\caption{Prior Knowledge Integrated DQN for Autonomous Guidewire Navigation}
\linespread{1}\selectfont
\begin{algorithmic}[1]
\STATE Initialize Q-network $Q_\theta$ and target network $Q_{\theta^-}$ with $\theta^- \leftarrow \theta$
\STATE Initialize replay buffer $\mathcal{D}$ with capacity $N$
\STATE Precompute distance matrix using Dijkstra's algorithm
\FOR{episode = 1 to $M$}
    \STATE Reset environment, obtain initial state $s_0$
    \STATE Initialize state history: $\text{history} \leftarrow [s_{\text{dummy}}, s_{\text{dummy}}, s_0]$
    \FOR{$t = 1$ to $T_{\text{max}}$}
        \STATE Select action $a_t$ using $\epsilon$-greedy policy:
        \STATE $a_t = \begin{cases}
        \text{random action} & \text{with probability } \epsilon \\
        \arg\max_a Q_\theta(s_t^{\text{stacked}}, a) & \text{otherwise}
        \end{cases}$
        \STATE Execute $a_t$, observe reward $r_t$, next state $s_{t+1}$
        \STATE Update stacked state: $s_{t+1}^{\text{stacked}} = [s_{t-1}^{\text{stacked}}[3:], s_{t+1}]$
        \STATE Store transition $(s_t^{\text{stacked}}, a_t, r_t, s_{t+1}^{\text{stacked}})$ in $\mathcal{D}$
        \IF{$|\mathcal{D}| > \text{batch size}$}
            \STATE Sample mini-batch from $\mathcal{D}$
            \STATE Compute targets: \\
            $y_i = r_i + \gamma \max_{a'} Q_{\theta^-}(s_{i+1}^{\text{stacked}}, a')$
            \STATE Update $\theta$ to minimize $(y_i - Q_\theta(s_i^{\text{stacked}}, a_i))^2$
        \ENDIF
        \STATE Update target network: $\theta^- \leftarrow \tau\theta + (1-\tau)\theta^-$
        \STATE Update exploration rate: $\epsilon \leftarrow \epsilon \cdot \gamma_{\epsilon}$
    \ENDFOR
\ENDFOR
\end{algorithmic}
\end{algorithm}

In Algorithm 1, the distance matrix is precomputed using Dijkstra's algorithm on the vessel centerline graph:
\begin{equation}
d_{\text{map}}(x,y) = \text{shortest path length from } (x,y) \text{ to target}
\end{equation}
This provides a more accurate navigation guidance compared to Euclidean distance, especially in tortuous vascular structures with multiple bifurcations.

\subsubsection{Reward Components}

The reward function is meticulously designed to provide hierarchical feedback that guides the agent through the complex navigation task while ensuring training stability. The complete reward structure incorporates multiple components that operate at different temporal scales and serve distinct pedagogical purposes.

\textbf{Integrated Reward Dynamics:} The complete reward function operates as an integrated system where components interact to create a coherent learning signal:

\begin{equation}
R_{\text{total}} = R_{\text{p}} + R_{\text{d}} + R_{\text{a}} + R_{\text{g}}
\end{equation}

The hierarchical design ensures appropriate reward assignment: the base penalty provides continuous efficiency pressure, distance feedback offers immediate directional guidance, achievement rewards recognize substantive progress, and the goal bonus provides clear terminal motivation. This multi-scale feedback structure enables the agent to learn both fine-grained maneuvering strategies and overall navigation objectives while maintaining the stability required for effective reinforcement learning in this safety-critical domain.

\textbf{Time Penalty:} Each step incurs a small but consistent penalty $R_\text{p}$. This design serves two critical functions: first, it encourages procedural efficiency by discouraging unnecessary deliberation or oscillatory behaviors; second, it establishes a baseline cost that the agent must overcome through meaningful progress, thereby promoting purposeful action selection. The modest magnitude ensures that this penalty provides gentle guidance without overwhelming the learning signal from more substantial achievement-based rewards.

\textbf{Distance Progression Feedback:} A fundamental component provides immediate directional guidance based on the agent's movement relative to the target:

\begin{equation}
R_{\text{d}} = 
\begin{cases}
+w_{\text{d}}, & \text{if } d_t < d_{t-1} \\
-w_{\text{d}}, & \text{if } d_t \geq d_{t-1}
\end{cases}
\end{equation}

This binary feedback mechanism offers clear, instantaneous assessment of each action's effectiveness in reducing the distance to the target. The symmetry between positive and negative feedback prevents reward asymmetry that could lead to risk-averse behaviors, while the modest magnitude prevents this component from dominating the overall reward signal.

\textbf{Progressive Achievement Reward:} To recognize and incentivize substantive navigation progress, the system implements a sophisticated achievement-based reward that activates when the agent establishes a new closest approach record:

\begin{equation}
R_{\text{a}} = 
\begin{cases}
\dfrac{d_{\text{initial}} - d_{\text{shortest}}}{d_{\text{initial}}}, & \text{if } d_t < d_{\text{shortest}} \text{ and } a_t \in \{0, 1\} \\
0, & \text{otherwise}
\end{cases}
\end{equation}

This component exhibits several sophisticated design characteristics:

\begin{itemize}
    \item \textbf{Action Specificity:} The reward is exclusively available for translational maneuvers (actions 0 and 1, corresponding to insertion/retraction operations), reflecting the clinical reality that meaningful distance reduction primarily occurs through axial movement rather than rotational adjustments.
    
    \item \textbf{Progressive Scaling:} The variable component $({d_{\text{initial}} - d_{\text{shortest}}})/{d_{\text{initial}}}$ creates a reward magnitude that is proportional to the overall progress achieved from the initial position. This design ensures that early, easy progress receives modest recognition, while later, more challenging advancements near the target receive appropriately scaled rewards, maintaining consistent learning incentives throughout the navigation trajectory.
    
    \item \textbf{Memory Mechanism:} The persistent tracking of $d_{\text{shortest}}$ ensures that the agent is rewarded for genuine progression milestones rather than transient improvements, effectively discouraging the exploitation of local minima or oscillatory patterns that might temporarily reduce distance without meaningful advancement.
\end{itemize}

\textbf{Goal Achievement Bonus:} A substantial sparse reward marks successful task completion, providing a clear terminal objective for the learning process:

\begin{equation}
R_{\text{g}}(s_t) =
\begin{cases}
+w_{\text{g}}, & \text{if } d_t \leq \varepsilon_{\text{goal}} \\
0, & \text{otherwise}
\end{cases}
\end{equation}

where $\varepsilon_{\text{goal}}$ defines the clinically acceptable target region. This generous bonus serves as the primary motivator for the agent to complete the navigation task, with the magnitude carefully calibrated to dominate the cumulative step penalties while remaining within numerically stable bounds for reinforcement learning algorithms.

The reward function's parameter values were carefully calibrated through extensive experimentation to provide clear learning signals while avoiding reward hacking, encouraging efficiency without promoting reckless behavior, and recognizing progress while maintaining numerical stability for gradient-based learning algorithms.

This meticulously designed reward function is a cornerstone of our autonomous navigation system, effectively translating surgical expertise and anatomical knowledge into a form that guides the DRL agent toward mastering the complex task of endovascular guidewire navigation.

\section{Experiments and Results}
In this section, we aim to systematically evaluate the performance of the proposed prior knowledge integrated DRL autonomous guidewire navigation system through comparative experiments and quantitative analyses.

\subsection{Experimental Setup}
\subsubsection{Simulation Environment and Vascular Model}
All experiments were conducted on a high-fidelity endovascular-intervention simulator built upon SOFA v24.12.00; the deep-reinforcement-learning algorithm and guidewire-motion controller were implemented in Python and executed on an Intel Core i9-14900K CPU together with an NVIDIA GeForce RTX 4070 Super GPU. The vascular model employed was a multi-bifurcation hepatic-vein phantom, as shown in Fig.~\ref{overall_system}. Physical parameters of the guidewire including flexural rigidity and friction coefficients were calibrated to replicate clinical instrumentation. The navigation objective was to steer the guidewire from a predefined entry port to a designated lesion target.

\subsubsection{Evaluation Metrics}
\begin{itemize}
  \item Success Rate (SR): The proportion of test episodes in which the guidewire successfully reaches the target region.

  \item Average Steps (AS): The mean number of steps required to successfully reach the target, reflecting navigation efficiency.

\end{itemize}

\subsubsection{Implementation Details}
All DRL algorithms were implemented in PyTorch. The hyperparameter Settings is as shown in Table~\ref{Hyperparameter Settings}.

\begin{table}[htbp]
\centering
\caption{Hyperparameter Settings.}
\label{Hyperparameter Settings}
\begin{tabular}{lcc}
\toprule
Parameter & Value \\
\midrule
Learning rate & 0.0005  \\
Discount factor ($\gamma$) & 0.99  \\
Replay buffer size & 10,240  \\
Batch size & 64  \\
Target update frequency & 1000 steps  \\
Total iteration times & 27000  \\
\bottomrule
\end{tabular}
\end{table}

\subsection{Comprehensive Performance Analysis and Discussion}
\label{sec:analysis}

To systematically evaluate the efficacy of the proposed prior knowledge integrated DRL framework, comparative experiments were conducted across three distinct configurations: (1) \textbf{Euclidean-Continuous}: A baseline employing a continuous state space and a Euclidean distance-based reward function; (2) \textbf{Dijkstra-Continuous}: An intermediate variant utilizing the proposed Dijkstra-based distance reward while retaining the continuous state space; and (3) \textbf{Dijkstra-Discretized (Ours)}: Our full model integrating the Dijkstra-based reward with the discretized and stacked state representation. This design allows for a granular ablation study, isolating the individual contributions of the reward metric and the state space design, and demonstrating their synergistic integration.

\subsubsection{Navigation Success and Operational Efficiency}
\label{subsec:success_efficiency}

The results summarized in Table~\ref{tab:performance_comparison}, unequivocally demonstrate the superior performance of our method under identical training iterations. The most striking evidence is the dramatic improvement in success rate achieved with the same computational cost.

\begin{table}[htbp]
  \centering
  \caption{Performance comparison of different algorithm configurations.}
  \label{tab:performance_comparison}
  \begin{tabular}{lcccccc}
    \toprule
    Method & Trials & Success Rate & Average Steps  \\
    \midrule
    Euclidean-Continuous   & 50 & 00\%   & -- \\
    Dijkstra-Continuous    & 50 & 98\%   & 64 \\
    Dijkstra-Discretized (Ours)   & 50 & 100\%  & 58 \\
    \bottomrule
  \end{tabular}
\end{table}

\begin{figure*}[htbp]
\centering
\includegraphics[width=1\linewidth]{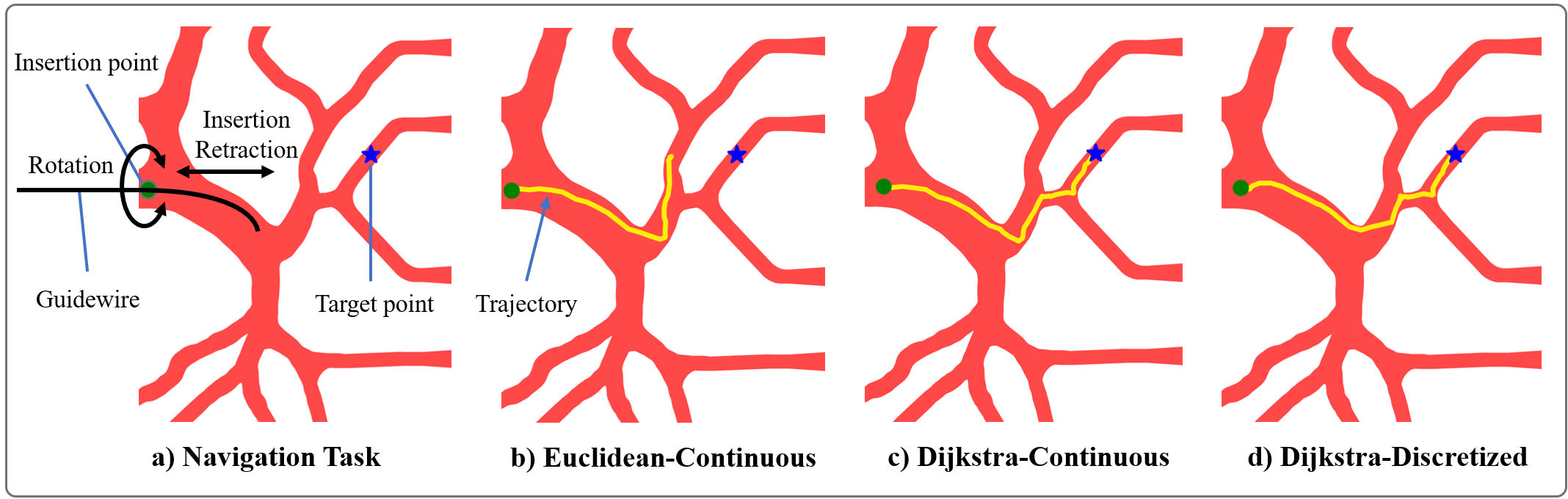}
\caption{Trajectory comparison. a) Illustrates the autonomous guidewire navigation setup, marking the initial point and the target point. b) The baseline method using Euclidean distance fails to reach the target, getting trapped in local optima due to misleading reward signals. c) By introducing the Dijkstra-based reward, the Dijkstra-Continuous method achieves 98\% success rate , showing immediate improvement over the Euclidean baseline. d) The proposed full model demonstrates rapid convergence, achieving 100\% success rate with the smoothest and most efficient trajectory.}
\label{fig:path}
\end{figure*}

Our full model, Dijkstra-Discretized, achieved a perfect 100\% success rate, while the Dijkstra-Continuous method reached 98\%. And the Euclidean-Continuous baseline failed to learn any viable policy. More importantly, our model demonstrated enhanced operational efficiency: the average number of steps required for successful navigation was 58, compared to 64 for the Dijkstra-Continuous method. This indicates that the discretized state representation not only facilitates learning reliability but also enables the refinement of more efficient policies. The higher average step count for the Dijkstra-Continuous method suggests a convergence to a sub-optimal policy, unable to further optimize for path optimality.

\subsubsection{Training Convergence Acceleration}
\label{subsec:convergence}

The superiority of our proposed framework is most evident in its dramatic acceleration of training convergence, a critical metric for assessing the practical viability of Deep Reinforcement Learning algorithms. The performance data at fixed iteration milestones, as detailed in Table~\ref{tab:performance_comparison}, provides compelling quantitative evidence of this advantage.

The superiority of our proposed framework is evident in its ability to converge to a highly reliable and efficient policy. As detailed in Table II, at 27,000 iterations, our model achieved 100\% success rate, indicating full convergence. Crucially, it also optimized its policy for efficiency, achieving the lowest average step count. In contrast, while the Dijkstra-Continuous method achieved a high success rate (98\%), its performance in terms of path optimality stagnated, with the average steps remaining higher. This indicates that the continuous state space presents a more challenging optimization landscape, leading to potentially less complete convergence. Such robust performance is paramount for the practical application of DRL in clinical settings, where extensive training cycles are prohibitive. The significantly reduced resource consumption underscores the practical viability and scalability of our integrated approach.

\subsubsection{Qualitative Behavior Analysis}
\label{subsec:qualitative}

The qualitative evidence, as illustrated by the trajectory analysis in Fig.~\ref{fig:path}, provides even deeper insights into the behavioral differences between the methods. At vascular bifurcations, the agent trained with the Euclidean-Continuous configuration exhibited pronounced hesitation and oscillatory behavior.

This counter-intuitive behavior is a direct consequence of the flawed Euclidean reward signal. In tortuous vasculature, the Euclidean distance can paradoxically increase as the agent moves down the correct but curved branch, effectively punishing correct actions and creating local optima that trap the agent. In contrast, both Dijkstra-based methods, guided by the monotonically decreasing gradient of the distance field, produced smooth, direct trajectories and confidently selected the correct branches. The trajectory of our full model was the most direct and stable, as shown in Fig.~\ref{fig:path}. This qualitative superiority proves that the Dijkstra-based metric provides a anatomically consistent and unambiguous learning signal, which is paramount for robust policy learning in complex geometries.

\subsubsection{Synergistic Integration}
\label{subsec:synergy}

The most significant performance leap is observed in the direct comparison between the Euclidean-Continuous baseline and our full Dijkstra-Discretized model. Our integrated framework achieves the highest success rate, the shortest operation time, and the fastest training convergence.

This demonstrates a powerful synergistic effect rather than a mere additive one. The Dijkstra-based reward ensures the agent learns the correct navigation strategy by providing a clinically-grounded learning signal. Simultaneously, the discretized state space enables the agent to learn this strategy efficiently and stably by reducing the complexity of the function approximation problem and mitigating partial observability. The combination of these two elements effectively addresses the two fundamental challenges of applying DRL to surgical navigation: inefficient reward shaping and the curse of dimensionality. The results validate the overall superiority and practical value of the proposed integrated framework, paving the way for more sample-efficient and robust autonomous navigation systems in vascular interventional surgery.

\section{Conclusions}

A novel DRL framework that integrates domain-specific prior knowledge is presented to overcome the fundamental limitations of autonomous guidewire navigation in vascular interventional surgery. Our investigation yields several key conclusions:

First, the choice of reward function metric is a critical determinant of success. Replacing the conventional Euclidean distance with a Dijkstra-based distance proved to be a effective advancement. This metric provides an anatomically meaningful and monotonically decreasing learning signal, which is essential for guiding the agent effectively through tortuous vascular pathways and preventing the oscillatory behaviors and local optima commonly induced by flawed Euclidean metrics.

Second, the design of the state space representation is equally crucial for training efficiency. Our introduction of a discretized and temporally stacked state space successfully addressed the high-dimensional continuous data. This design dramatically reduced the complexity of the function approximation problem, which was the key to achieving a remarkable reduction in the number of training episodes required for convergence, thereby significantly lowering the computational cost.

Finally, the synergistic integration of these two components—the anatomically accurate reward and the efficient state representation—constitutes the core of our contribution. This integration enables the learning process to be both well-guided and computationally tractable, leading to the emergence of robust, efficient, and clinically viable navigation policies.

In summary, this work provides a concrete and effective solution to the challenges of autonomous guidewire navigation in Vascular Interventional Surgeries. The proposed framework marks a step toward practical and efficient training of autonomous navigation systems, with the potential to alleviate surgeon burden and enhance the safety of endovascular procedures. Future work will focus on further validating the generalizability of the approach across more complex and diverse vascular anatomies and accelerating the transfer of policies learned in simulation to real-world robotic platforms.

\section*{Funding}
This work was supported by the High level of special funds under Grant G03034K003 from Southern University of Science and Technology, Shenzhen, China, in part by the Shenzhen “Medical Engineering Integration” Special Support Program Project under Grant F-2024-Z99-502365, Shenzhen, China.

\end{document}